\documentclass[letterpaper, 10 pt, journal, twoside]{ieeetran}
\IEEEoverridecommandlockouts                              % This command is only needed if 
\expandafter\let\csname equation*\endcsname\relax
\expandafter\let\csname endequation*\endcsname\relax

\title{Geometric Mechanics of Contact-Switching Systems}

\author{Hari Krishna Hari Prasad$^{1}$, Ross L. Hatton$^{2}$, and Kaushik Jayaram$^{1,*}$
\thanks{$^{1}$Animal Inspired Movement and Robotics Laboratory, Paul M. Rady Department of Mechanical Engineering, University of Colorado Boulder} 
\thanks{$^{2}$ Mechanical, Industrial and Manufacturing Engineering, Oregon State University} 
\thanks{$^{*}${For correspondence, \tt\footnotesize kaushik.jayaram@colorado.edu}}%
} % Use only for final RAL version.

\IEEEoverridecommandlockouts
\usepackage{mathnotation}
\usepackage{amsmath,amssymb,amsfonts}
\usepackage{tensor}
\usepackage{algorithmic}
\usepackage{graphicx}
\usepackage{graphics}
\graphicspath{{./Figures/Iter2/}}
\usepackage{textcomp}
\usepackage{xcolor}
\usepackage{gensymb}
\usepackage[utf8]{inputenc}
\usepackage{times}
\usepackage{mathtools}
\usepackage{subcaption}
\usepackage[separate-uncertainty=true,list-units=repeat,multi-part-units=single]{siunitx}
\usepackage{nicefrac}
\usepackage{multicol}
\usepackage{multirow}
\usepackage{array}
\usepackage{pgfplots} 
\usepackage{kantlipsum}
\usepackage{setstack}
\pgfplotsset{compat=1.14}
\usepackage{float}
\usepackage{stfloats}
\usepackage[flushleft]{threeparttable}
\usepackage{comment}
\usepackage{enumitem}
\usepackage{todonotes}

\newlength{\imagewidth}

\usepackage[numbers]{natbib}
\usepackage[pageanchor=true,plainpages=false, pdfpagelabels, bookmarks,bookmarksnumbered,hidelinks]{hyperref}

\usepackage{color} % For highlighting
\newcommand{\hari}[1]{\textcolor{black}{#1}}

\begin{document}
\maketitle

\vspace{-10pt}
\begin{abstract}
Discrete and periodic contact switching is a key characteristic of steady-state legged locomotion. This paper introduces a framework for modeling and analyzing this contact-switching behavior through the framework of geometric mechanics on a toy robot model that can make continuous limb swings and discrete contact switches. The kinematics of this model form a hybrid shape-space and by extending the generalized Stokes' theorem to compute discrete curvature functions called \textit{stratified panels}, we determine average locomotion generated by gaits spanning multiple contact modes. Using this tool, we also demonstrate the ability to optimize gaits based on the system's locomotion constraints and perform gait reduction on a complex gait spanning multiple contact modes to highlight the method's scalability to multilegged systems.
\end{abstract}

\IEEEpeerreviewmaketitle

\section{Introduction}
\label{sec:intro}

%\cite{} -- any citations for a fundamental Geometric Mechanics paper?
Animals and robots achieve locomotion by generating environment-relevant, periodic body deformations. Tools from the field of {\textit{geometric mechanics}} \cite{kelly1995geometric, oliva2004geometric} have been a popular choice for modeling these interactions and generating optimal gaits \cite{hatton2015nonconservativity,bittner2018geometrically} in a wide variety of biological and robophysical systems \cite{astley2020surprising,chong2022coordinatingtiny}. 
This framework relies on the notion that deformations within a system (i.e., shape changes) directly correlate to changes in its world position as often is the case with quasi-static locomotion with continuous environment-appendage contact. 
This relationship is mathematically described as \textit{the local connection} \cite{hatton2015nonconservativity} and has been successfully applied to 
{the study of various modes of limbless locomotion in drag-dominated environments such as during slithering \cite{astley2020surprising}, swimming \cite{jacobs2012geometric,hatton2013geometric}, and burrowing \cite{dai2016geometric} to name a few.} \par

\begin{figure}[tbp]
    \centering \includegraphics[width=0.49\textwidth]{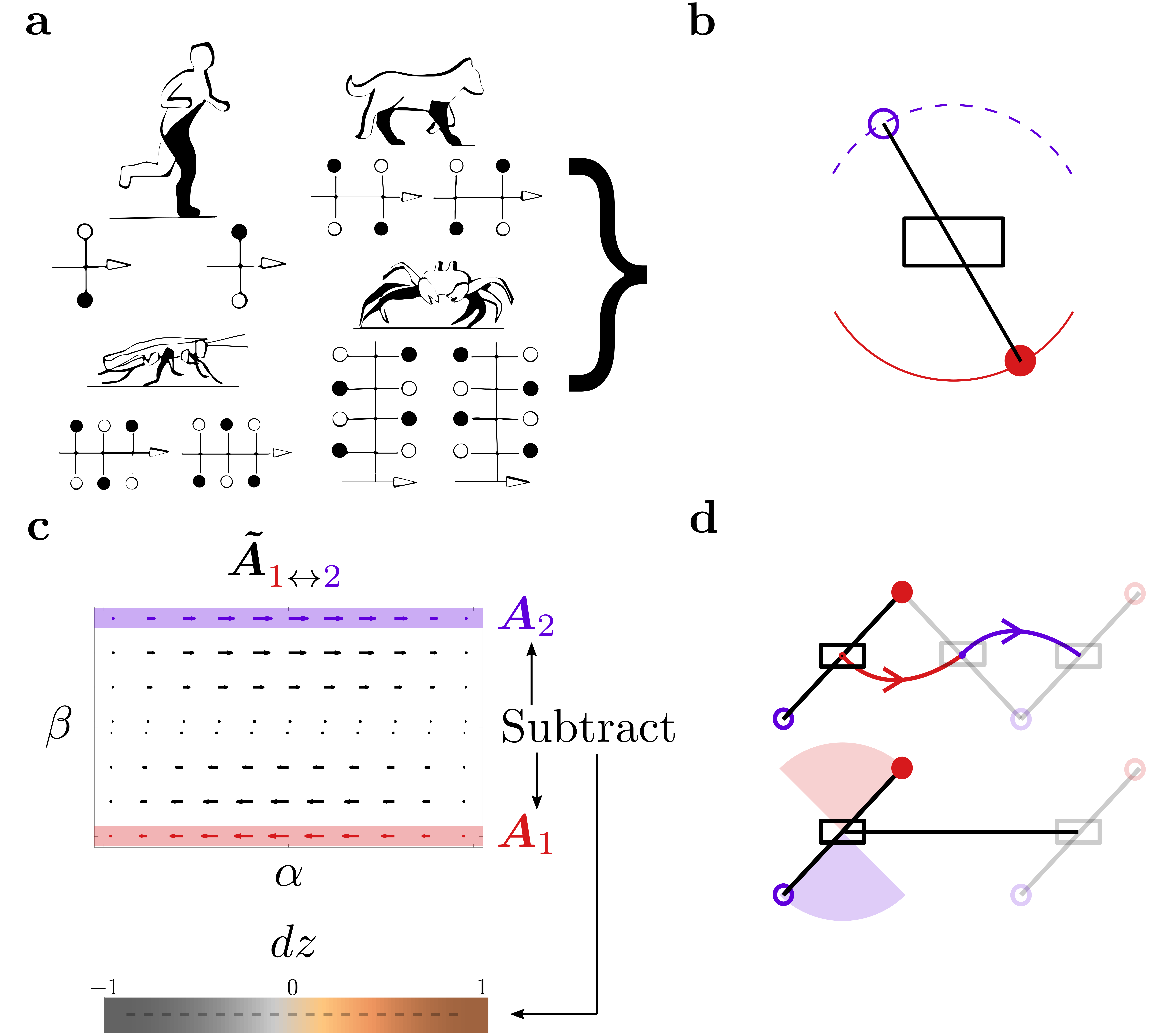}
    \caption{\hari{Overview of this work: \textbf{(a)} Hybrid dynamics of legged locomotion are a challenge for typical geometric mechanics tools. \textbf{(b)} The simplest illustrative formulation of a quasistatic legged locomotion system - a two-footed model. \textbf{(c)} The stratified locomotion panel framework presented in this work and its application for generating net body displacement over a gait cycle. \textbf{(d)} The piecewise holonomic trajectory (above) and the net displacement (below) of a two-footed contact switching system undergoing a single gait cycle.}}
    \label{fig:Overview}
\end{figure}

Recently, the geometric framework has been extended to analyze the locomotion of a variety of legged animals and robots. %\cite{zhao2022walking,chong2023frictionalswimming,chong2021coordbackbend,chong2019hierarchical,chong2022general,chong2021moving,chong2023optimizing}.
\hari{Zhao \textit{et al.} \cite{zhao2022walking} demonstrate that legged locomotion without slipping is always principally kinematic when the intermittent ground contact sequence is modeled as a function of body shape. As a result, the piece-wise holonomic nature of legged walking allows for cyclic shape changes to produce net locomotion similar to slithering and Stokesian swimming.}
However, as noted in \cite{chong2021coordbackbend}, ``The challenges of extending geometric mechanics to quadrupedal systems lie in the fact that these systems periodically make and break contact with the environment." 
These discrete and periodic appendage-environment interactions result in multiple local connections which distinguish it from the other modalities discussed above and limit the direct application of the established geometric mechanics framework.

To overcome the limitation imposed by contact-switching, one approach recasts the undulatory locomotion with leg retraction and protraction as a fluid-like problem with the nonlinearities of foot–ground interactions leading to acquired drag anisotropy and thus models multilegged locomotion as frictional swimming using existing tools \cite{chong2023frictionalswimming}.
In an alternate approach, Chong \textit{et al.} \cite{chong2021coordbackbend} use biological observations to prescribe a leg-contact sequence a priori utilizing a single continuous phase variable. A similar approach prescribes a second continuous phase variable to model body undulations. By coupling them together, these phases represent continuous shapes that form a single toroidal shape-space amenable to the geometric framework and capture the discontinuous aspect of legged locomotion through the prescribed leg-contact trajectory \cite{chong2019hierarchical,chong2022general}. 
\hari{Extending this approach, recent works from Chong \textit{et al.} \cite{chong2021moving, chong2023optimizing} combine multiple local connections in multi-link limbless and legged robots into a two-dimensional form by partitioning a single locomotion period into empirically observed contact regions using appropriate shape basis functions.
This allows them to compute displacement strengths from the discrete difference between system constraints in adjacent curvature-free modes, and optimize gait sequences by determining transitioning points that maximize net motion accrued in each contact state. 
However, this method is limited as the authors point out \cite{chong2021moving}, ``Simply choosing the local optima as the transitional points can lead to problems such as self-intersection gait paths or gait paths with extremely large perimeters”.
}

\hari{To address these limitations, we propose an alternate well-conditioned formulation for optimizing gaits in hybrid systems utilizing the \textit{chain of forms} approach \cite{hatton2022geometry} - a unified framework for integrating differential forms that mix continuous curvature with sharp corners, similar to how proper functions and Dirac delta functions can be meaningfully integrated together.} \par
{Specifically, to take advantage of this framework, }we model the contact-switching nature of kinematic legged locomotion using a hybrid shape-space consisting of both \textit{discrete} and continuous shape dimensions. For such a hybrid system, we show that the local curvature undergoes a discrete reduction into \textit{stratified panels} motivated by the nomenclature from non-linear geometric control \cite{goodwine2002motionkinstrat}. 
\hari{Thus, the key advantage of this formulation is that it only relies on contact-switching rather than the exact sequence or specific model of ground contact and thereby enables broader generalization. 
For example, unlike the above-mentioned approaches, our framework can be readily extended to analyze the locomotion of robots operating under constraints different from biological systems such as those able to choose arbitrary foot timing patterns or have the ability to modulate contact using adhesion mechanisms \cite{de2018inverted}.
%We wish to further highlight that our focus in this work is to analyze hybrid systems from a geometric perspective and therefore, our formulation deliberately deemphasizes the exact contact model description as is not important to model overall locomotion.
} 

\hari{To illustrate our methodology}, we use a two-footed toy model inspired by peg-legged walkers \cite{ruina1998nonholonomic} {to develop and present a new geometric mechanics framework relevant for gait generation and analysis of contact-switching systems in {\S \ref{sec:dual_cont}}. 
\hari{Extensions of this no-slip approach to model contact-switching systems with slipping or specific ground interactions are outlined at the end of this section for the sake of completeness while not specifically being the focus of this manuscript.} 
We then describe our proposed {stratified panel} method for drawing gaits and compute optimal gaits subject to constraints (costs) associated with leg swinging and contact switching in {\S \ref{sec:gait}}. 
\hari{We define our objective function for gait optimization as the ratio of
{the displacement accrued relative to the system's effort} over a locomotion cycle where both quantities are differential forms. The path cost metric (numerator) is obtained as a surface integral of the \emph{two-form} curvature functions, while the shape change cost (denominator) is calculated through a closed line integral of a \emph{quadratic form}. 
Thus, by leveraging the \textit{chain of forms} approach, we avoid the ill-conditioned problem formulation noted during previous gait optimization methods \cite{chong2021moving}.}
We also demonstrate the scalability of this approach for modeling complex gaits spanning multiple contact states in {\S \ref{subsec:three_cont}} by introducing a generalized gait coordinate system and illustrating the same with a three-contact system example. 
Finally, we discuss the broader applicability of this work in the context of other state-of-the-art approaches and conclude by hinting at exciting future directions in {\S \ref{sec:disc}}. The code repository for this work can be found here: \url{https://github.com/Animal-Inspired-Motion-And-Robotics-Lab/Paper-Geometric-Mechanics-of-Contact-Switching-Systems}.

\section{Two-foot, Contact-switching Locomotion}
\label{sec:dual_cont}

As the first step towards describing the geometric mechanics-based, stratified panel framework for quasistatic legged locomotion, we consider a two-footed toy robot model as the simplest instantiation of contact-switching walking (Fig. \ref{fig:Overview}).
\hari{We additionally note that our choice to focus on legged locomotion without slipping in this manuscript enables us to generate a minimal \emph{template-style} \cite{full1999templates} locomotion model which is sufficient to model the contact-switching mechanics of walking and explore geometric properties of periodic actuation in a hybrid shape-space.}

\vspace{-10pt}
\subsection{System Description}
\label{subsec:sys_des}

Our contact-switching model {illustrated} in Fig. \ref{fig:2c_model_ss_interp_strata}(a), is a planar system with two legs of length $r$ constrained diametrically opposite to each other and attached to the system's body frame, $g_b \, = \, (x, y, \theta) \in \mathit{SE}(2)$ \hari{(group of planar rigid-body motions)}. %concentric
For simplicity of analysis, we constrain the body to {move with only} translations and no rotations ($g_b^\theta=0$).\footnote{In a more physically realistic scenario, one could instead model the feet and body as each being subject to resistive friction forces, which would result in a model with the same structure but with more complicated expressions in the constraints as demonstrated in prior work \cite{chong2019hierarchical,hatton2017Riemannian}.}The robot is driven by a single rate-limited servo rotor controlling the leg angle, $\alpha$, {relative to the $g_b^x$ direction} with symmetric bounds {($|\alpha_\text{max}| = \frac{\pi}{2}$)}.

\begin{figure}[hbtp!]
    \centering \includegraphics[width=0.45\textwidth]{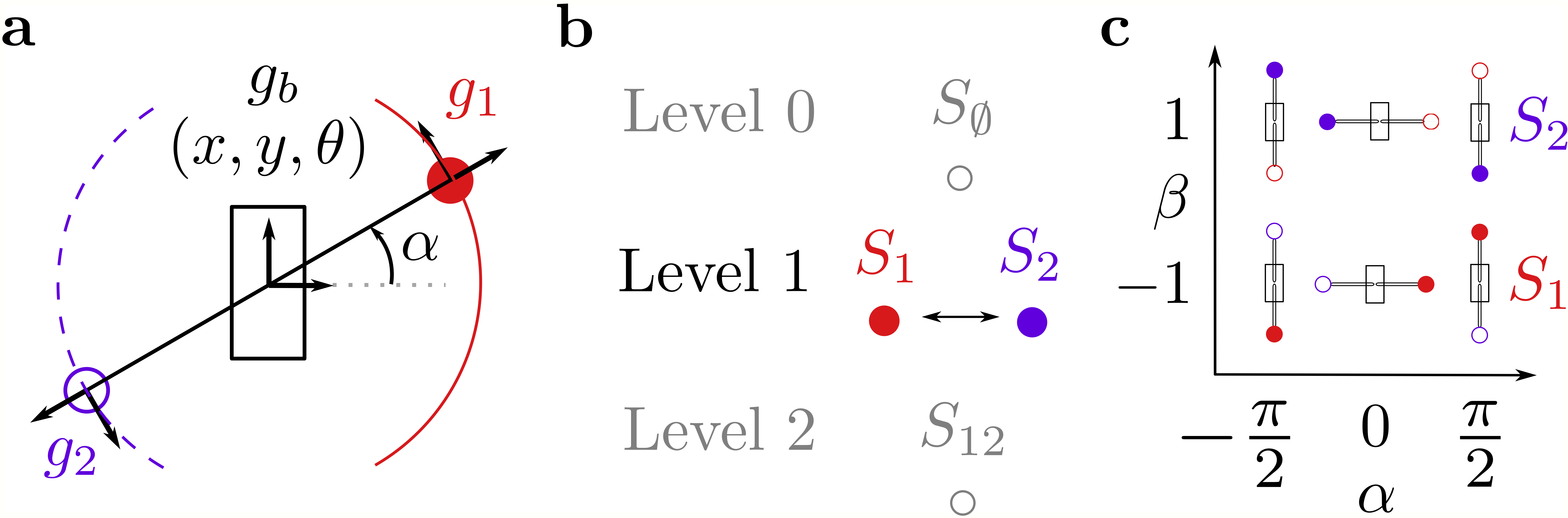}
    \caption{(\textbf{a}) A two-footed, contact-switching model moving its diametrically-coupled legs $g_1$ and $g_2$ in its body frame $g_b$ with leg angles $\alpha$ and $\pi + \alpha$ respectively. The pinned right foot-tip is highlighted and its trajectory is shown with a solid trace. (\textbf{b}) The locomotion submanifolds arising from different combinations of contacting legs are organized into levels based on the number of legs pinned at any given time. Submanifolds, $S_{\emptyset}$ and $S_{12}$ are degenerate cases that produce no motion. (\textbf{c}) \hari{The system's hybrid shape-space consists of a continuous rotor angle ($\alpha$), and a discrete foot contact state ($\beta$).} Multiple body configurations within the shape-space are illustrated.}
    \label{fig:2c_model_ss_interp_strata}
\end{figure}

To signify a foot making and breaking contact with its environment, we define a binary contact variable for each foot, $c_i \in \{0,1\}$. 
The state $c_i=1$ represents the robot maintaining contact with the ground via a freely rotating pin joint at its {$i$th} foot (also referred to as the stance phase during legged locomotion).
In contrast, the state $c_i=0$ represents the {$i$th} foot's aerial (or frictionless sliding or swing) phase which produces no system motion.
Furthermore, by following the previously established conventions for multi-legged robots \cite{goodwine2002motionkinstrat}, the system's current locomotion state {(or submanifold, \textit{S})} is denoted as $S_{ij\ldots}$.
%\footnote{The subscript includes an ascending-ordered list of every foot in contact and this method helps avoid permutations.}.

\vspace{-10pt}
\subsection{Kinematic Reconstruction using the Geometric Mechanics Framework during locomotion without slipping}
\label{subsec:kin_recon}
To compute the kinematics of the system by determining its \textit{ local connection}, we parameterize the
robot using $SE(2)$ frames as: $g_b$ at the body center, $g_i$ at the {$i$th} leg's foot-tip and the corresponding leg angle $\alpha_i$ (sum of the rotor angle, $\alpha$, and a offset angle of the leg). 
We then compute the Jacobian relating leg velocity, $\groupderiv{g}_i$ to the body velocity, $\groupderiv{g}_b$ and the rotor velocity, $\dot{\alpha}$ as:
\begin{align}
\label{eq:jacobian_2Ci}
  \left(\begin{array}{c}
    \groupderiv{g}_i^x \vspace{1mm}\\
    \groupderiv{g}_i^y \vspace{1mm}\\
    \groupderiv{g}_i^\theta
    \end{array}\right) = \left(\begin{array}{cccc}
                                            \hphantom{-}\cos \left(\alpha_i \right) & \sin \left(\alpha_i \right) & 0 & 0\\
                                                    -\sin \left(\alpha_i \right) & \cos \left(\alpha_i \right) & r & r\\
                                                    0 & 0 & 1 & 1
                                                    \end{array}\right) \left(\begin{array}{c}
                                                                        \groupderiv{g}_b^x \vspace{1mm}\\
                                                                        \groupderiv{g}_b^y \vspace{1mm}\\
                                                                        \groupderiv{g}_b^{\theta} \vspace{1mm}\\
                                                                        \dot{\alpha}
                                                                        \end{array}\right)
\end{align}

We apply the no-slip condition{ \cite{ruina1998nonholonomic} at each foot contact (i.e., $\groupderiv{g}_i^x=\groupderiv{g}_i^y=0$ and ignore $\groupderiv{g}_i^{\theta}$
%\footnote{\hari{Setting the translational velocities to zero provides us with two constraint equations and three ways to move in $\mathit{SE}(2)$ for the single-foot, nonslip contact condition. This means more degrees of freedom than constraints and is the reason we choose to work with just planar translations (the translation sub-group of $\mathit{SE}(2)$).}})
to obtain the Pfaffian constraint matrix during $S_{i}$ submanifold locomotion as:
\begin{align}
    & \left(\begin{array}{c}
    0 \\
    0 \\
    \end{array}\right) = \left(\begin{array}{cccc}
    \hphantom{-}\cos \left(\alpha_i \right) & \sin \left(\alpha_i \right) & 0\\
    -\sin \left(\alpha_i \right) & \cos \left(\alpha_i \right) & r
    \end{array}\right) \left(\begin{array}{c}
    \groupderiv{g}_b^x \vspace{1mm} \\
    \groupderiv{g}_b^y \vspace{1mm} \\
    \dot{\alpha}
    \end{array}\right) \label{eq:pfaffR_mod_2C_2}
\end{align}

Body motion can occur only if a single foot is hinged to the ground (i.e. either $c_1=1$ during $S_{1}$ or $c_2=1$ during $S_{2}$) but not both (i.e. $c_1=c_2=1$ during $S_{12}$). Hence, submanifolds that involve only one leg in contact are considered in our analysis as shown in Fig. \ref{fig:2c_model_ss_interp_strata}(b). Reorganizing {\eqref{eq:pfaffR_mod_2C_2}}, we obtain the kinematic reconstruction equation \cite{hatton2015nonconservativity} for the two-footed system during $S_{i}$ in the form $\groupderiv{g} = -\bf{A(\alpha)}\dot{\alpha}$ {as}:
\begin{align}
    & \left(\begin{array}{c}
    \groupderiv{g}_b^x  \vspace{1mm}\\
    \groupderiv{g}_b^y  \\
    \end{array}\right) = \, -\left(\begin{array}{c}
                -r\,\sin \left(\alpha_i \right)\\
                \hphantom{-}r\,\cos \left(\alpha_i \right)
                \end{array}\right)
    \dot{\alpha}
     \label{eq:kin_recon}
\end{align}
Substituting the leg angles, $\alpha_1=\alpha$ and $\alpha_2=\pi + \alpha$ (due to the mechanical coupling), we obtain two local connections, $\boldsymbol{A}_{1}$ and $\boldsymbol{A}_{2}$, corresponding to $S_{1}$ and $S_{2}$ respectively as:
\begin{align} 
    \boldsymbol{A}_{1} (\alpha) = \left(\begin{array}{c}
                -r\,\sin \left(\alpha \right)\\
                \hphantom{-}r\,\cos \left(\alpha \right)
                \end{array}\right);
    \boldsymbol{A}_{2} (\alpha) = \left(\begin{array}{c}
                \hphantom{-}r\,\sin \left(\alpha \right)\\
                -r\,\cos \left(\alpha \right)
                \end{array}\right) \label{eq:A_12}
\end{align}
Thus, to fully specify our two-footed system, we require two distinct connections, one for each motion-viable submanifold. This stratified formulation of the contact-switching kinematics follows directly from geometric control of legged locomotion \cite{goodwine2002motionkinstrat} and represents a distinct departure from previous geometric modeling of multilegged walking \cite{chong2022coordinatingtiny, chong2021coordbackbend, chong2019hierarchical} or limbless slithering \cite{hatton2013geometric, hatton2013geometric2} where a single local connection was sufficient. Therefore, the pin constraint resulting from hybrid motion makes connection vector fields \textit{conservative} for any number of dimensions, and \textit{singular} in shape directions unrelated to the pinned leg \hari{supporting the claim that legged locomotion without slipping is principally kinematic \cite{zhao2022walking}.} 

\hari{For classes of quasistatic locomotion systems undergoing slipping, the corresponding local connections can be obtained by balancing the forces on the system computed using the resistive force theory (RFT) approach \cite{gray1955propulsion} instead of applying the no-slip condition \cite{ruina1998nonholonomic} as indicated above. 
We refer the readers to published recent literature on geometric modeling of slipping locomotion on hard ground using an ansatz friction model \cite{zhao2021locomotion, wu2019coulomb} and  on fluid-like substrates \cite{ chong2021coordbackbend, chong2023frictionalswimming, chong2021moving} using granular RFT \cite{zhang2014effectiveness}  for details of applying these methods to obtain descriptions similar to \eqref{eq:jacobian_2Ci} and \eqref{eq:kin_recon}.
However, we note that these approaches, typically require additional information about the contact models, friction coefficients, leg sequencing, etc. to accurately estimate motion, drawing attention away from the contact-switching nature of such locomotion and is therefore, currently outside the scope of this work.
}

\vspace{-5pt}
\subsection{Contact-switching Interpolation Extension}
\label{subsec:trad_meth}

In order to leverage existing tools to analyze systems with stratified kinematics, we introduce a single, switching-mode (or hybrid) local connection ({$\boldsymbol{\Tilde{A}}_{i \leftrightarrow j}$}) which combines multiple distinct local connections ($\boldsymbol{A}_{i}$ and $\boldsymbol{A}_{j}$). 
To accomplish this in our case, we introduce a discrete \textit{contact} shape variable, $\beta \in \{-1,1\} \cap \mathbb{Z}$ and an interpolated contact shape, $\Tilde{\beta} \in [-1,1] \cap \mathbb{R}$ to smoothly interpolate between the two vector fields. To complete the formulation, we define a continuous version,\footnote{A continuous version of the hybrid local connection is introduced to formally simplify Stokes' surface integration to a discrete case in \S \ref{subsec:panel_meth}} $\Tilde{c_i}$, of our previous defined contact states, $c_i$, for each leg (shown in Fig. \ref{fig:vecF_CCF_panels_2C}a) as:
\begin{align}
    \Tilde{c_1} (\Tilde{\beta}) = \frac{1 + \cos{\frac{\pi}{2} (\Tilde{\beta} + 1)}}{2},
    \Tilde{c_2} (\Tilde{\beta}) = \frac{1 - \cos{\frac{\pi}{2} (\Tilde{\beta} + 1)}}{2} \label{eq:sineInterp_2C_ab}
\end{align}

As the next step, we define the switching-mode local connection, $\boldsymbol{\Tilde{A}}_{1 \leftrightarrow 2}$ (shorthanded hereafter as $\boldsymbol{\Tilde{A}}$) as:
\begin{equation} \label{eq:InterpConn}
    \boldsymbol{\Tilde{A}}_{1 \leftrightarrow 2} (\alpha,\Tilde{\beta}) = \Tilde{c_1} (\Tilde{\beta}) \, \boldsymbol{A}_{1} (\alpha) + \Tilde{c_2} (\Tilde{\beta}) \, \boldsymbol{A}_{2} (\alpha)
\end{equation}
The discrete version of the switching-mode connection is a piecewise function (either $\boldsymbol{A}_{1}$ or $\boldsymbol{A}_{2}$) based on the discrete contact shape ($\beta$) as highlighted in Fig. \ref{fig:vecF_CCF_panels_2C}(b).

\begin{figure} [btp]
    \centering \includegraphics[width=0.49\textwidth]{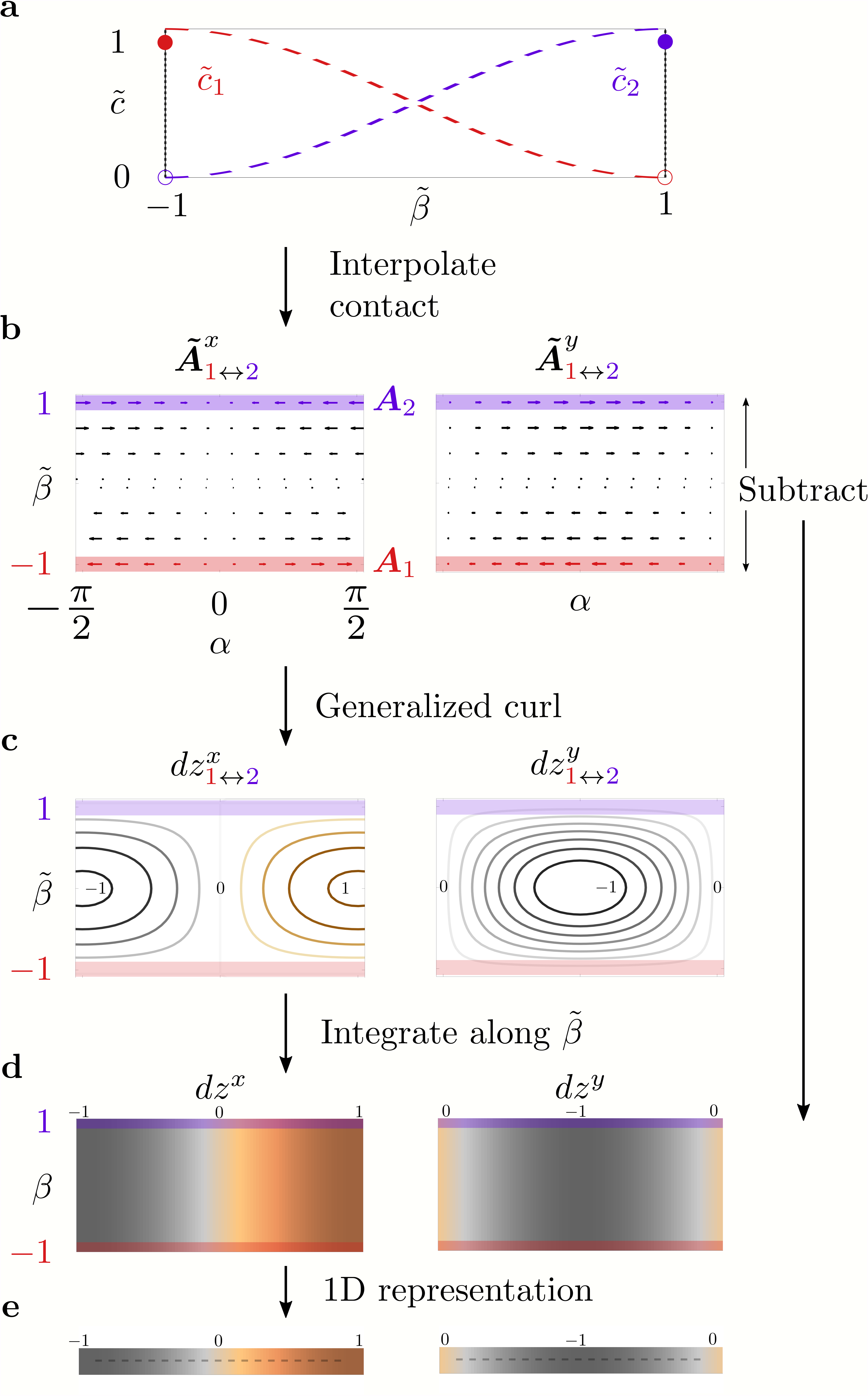}
    \caption{(\textbf{a}) The interpolated contact-state functions, $\Tilde{c}_1$ and $\Tilde{c}_2$ aid in formulating a continuously changing, locomotion submanifold, $S_{1 \leftrightarrow 2}$ and the corresponding (\textbf{b}) interpolated connection vector fields (each submanifold is highlighted); followed by (\textbf{c}) constraint curvature functions in translational coordinates obtained from Stokes' surface integral, (\textbf{d}) the stratified panels that encode displacement accrued from infinitesimal gaits at each $\alpha$ (vertical strips), and (\textbf{e}) finally, a collapsed one-dimensional representation of the stratified panel.}
    \label{fig:vecF_CCF_panels_2C}
\end{figure}

\vspace{-5pt}
\subsection{Stratified Panel Method Simplification - The Switching Flux}
\label{subsec:panel_meth}

Using the switching-mode local connection {\eqref{eq:InterpConn}}, we compute the system's displacement, $z_\phi$, over a given gait cycle, $\phi \in \Phi$, by a generalized Stokes' Theorem based surface integral \cite{hatton2015nonconservativity} as:
\begin{align}
    & {z}_\phi = \iint_{\phi_a} -\left( \frac{\partial{\boldsymbol{\Tilde{A}}^{(2)}}}{\partial{\alpha}} - \frac{\partial{\boldsymbol{\Tilde{A}}^{(1)}}}{\partial{\Tilde{\beta}}} \right) \,d\alpha\,d\Tilde{\beta} \label{eq:CCFeqn_hybrid2C_a}
\end{align}
\noindent where $\phi_a$ is the signed region of the shape-space enclosed by $\phi$, and $\boldsymbol{\Tilde{A}}^{(1)}$ and $\boldsymbol{\Tilde{A}}^{(2)}$ are the two columns in $\boldsymbol{\Tilde{A}}$. 
{Since rotations \eqref{eq:pfaffR_mod_2C_2} are disabled in this system, \eqref{eq:CCFeqn_hybrid2C_a} does not include the first-order Lie bracket.} The associated integrands (Fig. \ref{fig:vecF_CCF_panels_2C}c) called \textit{constraint curvature functions} (CCFs \hari{\cite{hatton2015nonconservativity}}, or height functions) quantify the displacement strength {over an infinitesimal cycle at} each point in the shape space. 
Since switching contact does not cause motion, mathematically, $\boldsymbol{\Tilde{A}}^{(2)}$ is a null vector, and enables us} to obtain a simpler formulation of the CCFs specific to hybrid locomotion systems called \textit{stratified panels} (Figs. \ref{fig:vecF_CCF_panels_2C}d and \ref{fig:vecF_CCF_panels_2C}e) as: 
\begin{align}
    {z}_\phi & = \iint_{{}^v\phi_a} \frac{\partial{\boldsymbol{\Tilde{A}}^{(1)}}}{\partial{\Tilde{\beta}}} \,d\Tilde{\beta}\,d\alpha 
     = \int_{\alpha^{-}}^{\alpha^{+}} \left[ \boldsymbol{\Tilde{A}}^{(1)} \right]_{\Tilde{\beta} = -1}^{1} \,d\alpha \label{eq:zphi_flux_a} \\
    & = \int_{\alpha^{-}}^{\alpha^{+}} -\left( \boldsymbol{A}_{1} - \boldsymbol{A}_{2} \right) \,d\alpha = \int_{\alpha^{-}}^{\alpha^{+}} \,dz \,\,d\alpha \label{eq:zphi_flux_b}
\end{align}

The stratified panel ($dz$) is, therefore, the flux line along $\alpha$ representing the infinitesimal switching flux and the displacement obtained by an infinitesimal gait centered about $\alpha$ between submanifolds $S_{i}$ and $S_{j}$ is simply the stratified panel (i.e. the difference between their local connections, -($\boldsymbol{A}_{i}$-$\boldsymbol{A}_{j}$)) multiplied by the infinitesimal leg excursion.
{Furthermore, \eqref{eq:zphi_flux_b} indicates that the choice of a contact interpolation function, $\Tilde{c_i}$ in \eqref{eq:sineInterp_2C_ab}, provided it is at least differentiable once, {does not} affect the stratified panel and consequently the resultant displacement.}

\section{Stratified Panel Method for Motion Estimation and Optimal Gait Generation}
\label{sec:gait}
{So far,} we have demonstrated that the integral around a gait loop with one continuous variable and one switched variable is equal to the integral along the corresponding stratified panel. We use this principle to illustrate our procedure for motion estimation and gait optimization in the following subsections.

\vspace{-10pt}
\subsection{Motion Estimation}
\label{subsec:gait_syn}
We choose a sample discrete gait (parameterized by time, $\tau$) switching between foot extremal positions ($\alpha^-$ and $\alpha^+$, Fig. \ref{fig:discrete_gait_2C}a). 
To compute motion over a cycle, this gait is transposed onto the system's stratified panels \eqref{eq:zphi_flux_b} along the x ($dz^x$) and y ($dz^y$) coordinate axes respectively as shown in Fig. \ref{fig:discrete_gait_2C}b. We observe that the $dz^{x}$ stratified panel is anti-symmetric about $\alpha = 0$, indicating that no net motion is accrued along that direction ($z_\phi^x=0$) for symmetric gaits. On the other hand, the symmetric $dz^{y}$ stratified panel indicates equal (due to the mechanical coupling) displacement in the y-direction is obtained during each contact phase. The net displacement (Fig. \ref{fig:discrete_gait_2C}c) accrued over the above gait-cycle, $z_\phi$, is simply the integration of the stratified panels along $\alpha$ and aligns with the robot center's trajectory (Fig. \ref{fig:discrete_gait_2C}d) computed using the boundary integral \cite{hatton2015nonconservativity}.

\begin{figure} [hbtp!]
    \centering \includegraphics[width=0.49\textwidth]{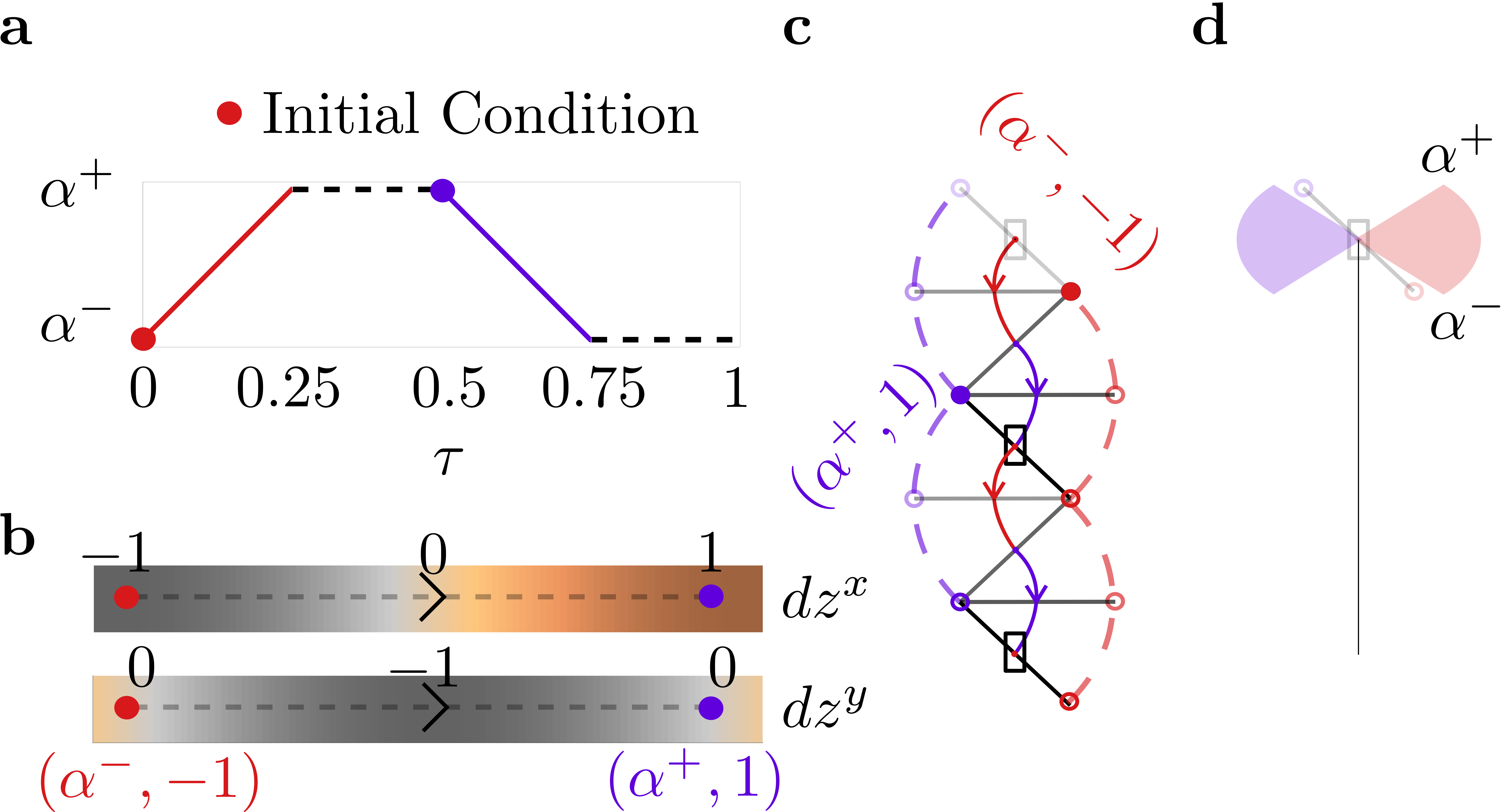}
    \caption{(\textbf{a}) The rotor angle shape change in a discrete gait as function of the time period $\tau$ {(highlighted regions denote the active contact state)}, \hari{(\textbf{b}) $x$ and $y$-displacement stratified panels are embedded} with a discrete gait that swings each leg from $\alpha^-$ to $\alpha^+$ in each contact state, (\textbf{c}) the corresponding trajectory of the system in $\mathit{SE}(2)$ obtained by a boundary integral, and (\textbf{d}) the net displacement is computed by integrating along the stratified panel from $\alpha^-$ to $\alpha^+$ {and is plotted as a vertical line. The limb excursions in this gait are shown as circular sectors for each limb (colored to denote each level-one, contact state) at the initial $\mathit{SE}(2)$ position of the system}.}
    \label{fig:discrete_gait_2C}
\end{figure}

\vspace{-15pt}
\subsection{Optimal Gait Generation}
\label{subsec:gait_opt}
%\cite{}
One of the strengths of the geometric approach for gait synthesis is the ability to generate {optimal} gaits subject to locomotion constraints typically formulated as cost functions involving performance metrics such as maximizing forward speed, minimizing the cost of transport, etc. To illustrate the gait optimization procedure, we choose locomotion effectiveness ($E_\phi^y$) as the cost function to be maximized and define it as the ratio of the $\mathit{SE}(2)$ displacement accrued by the robot's center (${z}_\phi$) to the benefit function associated with changing the shape-space ($J_\phi$) over a single gait cycle. \par
As discussed in \S{\ref{subsec:trad_meth}}, the displacement per cycle can be found from the surface integral of the constraint curvature (including the line and point integrals of the ``creases" and corners induced by discrete changes in the contact state). We take the cost as the time taken to complete a gait cycle at normalized instantaneous effort, which under our model of a rate-limited servo corresponds to the pathlength of the cycle in the space of joint angles.\footnote{For more detailed physical models, the pathlength calculation can be weighted with a quadratic form/Riemannian metric to reflect configuration-dependent relationships between motor speed and motor effort \cite{ramasamy2019draggeometry, hatton2022geometry}.} \par
For our system operated by a constant rate servo and translating along the Y direction, locomotion effectiveness ($E_\phi^y$) is a measure of \emph{average speed} dependent on leg speed ($s$), leg amplitude ($\hat{\alpha}$), and foot contact-switching time ($T_\beta$) as: 
\begin{equation} \label{eq:eff_2C_y}
    E_\phi^y = \frac{z_\phi^y}{J_\phi} = \frac{z_\phi^y}{T_\phi} = \frac{ 2r \left[ \sin{\alpha} \right]_{\alpha^-}^{\alpha^+} } { 2 s^{-1} \left[ \alpha \right]_{\alpha^-}^{\alpha^+} + 2T_\beta } = \frac{r \sin{\hat{\alpha}} } { T_\alpha + T_\beta }
\end{equation}
 
For ease of calculations, we assume the servo motor speed is fixed at half a revolution per second and the leg length $r$ is 1 unit. Fig. \ref{fig:eff_swing_2C} depicts the optimal swing amplitude solution ($\hat{\alpha}^*$) that maximizes locomotion effectiveness as a function of switch time ratios ($\frac{T_{\alpha}}{T_{\beta}}$) to validate our intuition about the behavior of rate-limited, contact-switching systems --- if the system is capable of relatively fast contact switches over leg swings (high $\frac{T_{\alpha}}{T_{\beta}}$), then it is preferable to make many small leg excursions (smaller $\hat{\alpha}^*$)  for maximizing effectiveness; else it is better to make few large leg swings (larger $\hat{\alpha}^*$).

\begin{figure} [hbtp!]
    \centering \includegraphics[width=0.49\textwidth]{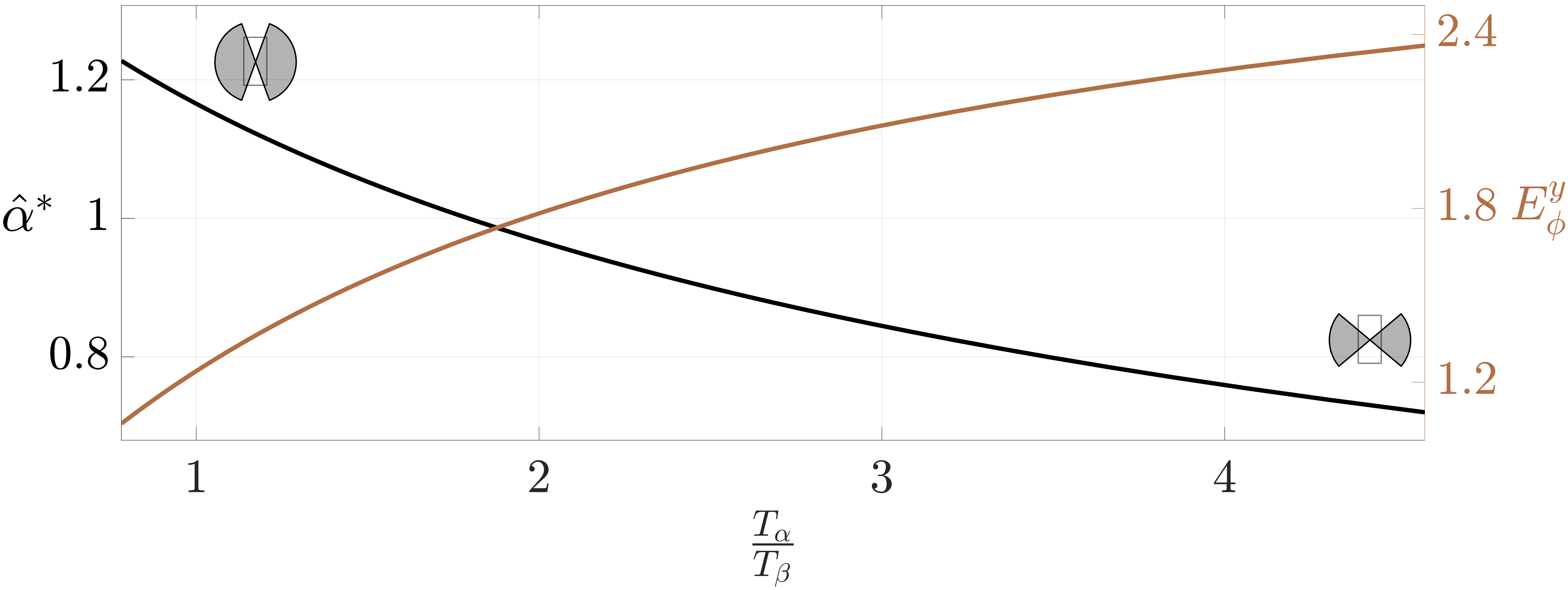}
    \caption{{The optimal swing amplitude $\hat{\alpha}^{*}$ and the optimal speed in y-direction $E_\phi^y$ for the two-footed, contact switching system are plotted as functions of the ratio between constant leg swing and contact switch times, $\frac{T_\alpha}{T_\beta}$. The optimal swing amplitudes of the system are shown as mini leg excursion insets at extremal values of $\frac{T_\alpha}{T_\beta}$.}}
    \label{fig:eff_swing_2C}
\end{figure}

\vspace{-5mm}
\section{Generalization to systems switching between multiple (three and above) contacts}
\label{sec:three_cont_gen}

In this section, we extend our results so far to contact-switching, multi-footed systems and describe a procedure for reducing the resulting complex gaits spanning multiple submanifolds in a high dimensional shape space using stratified panels. 

\vspace{-10pt}
\subsection{Three-footed, contact switching system}
\label{subsec:three_cont}
Specifically, a three-footed robot is chosen as an illustrative example as it increases the number of locomotion submanifolds by one relative to the two-footed robot and introduces three stratified panels as there are ways to choose two contact states to switch between. 
Similar to our previous system (Sec. \ref{subsec:sys_des}), the three-footed robot is a planar system with three legs of length $r$ attached to the system's body frame and constrained to be symmetrically distributed around the body (i.e., $120\degree$ offset between each leg). {All previous assumptions carry over and enable only body translations produced by leg swings in level one submanifolds, $S_{1}$, $S_{2}$, and $S_{3}$.} 
Furthermore, we assume switching between any two submanifolds accrues the same cost. Therefore, the natural embedding of the three-footed system in $\mathbb{R}^{3}$ resembles an equilateral, triangular prism with the rectangular faces encoding the stratified panels between two submanifold edges \hari{(Fig. \ref{fig:3C})}.

\begin{figure*}[tbp!]
    \centering \includegraphics[width=\textwidth]{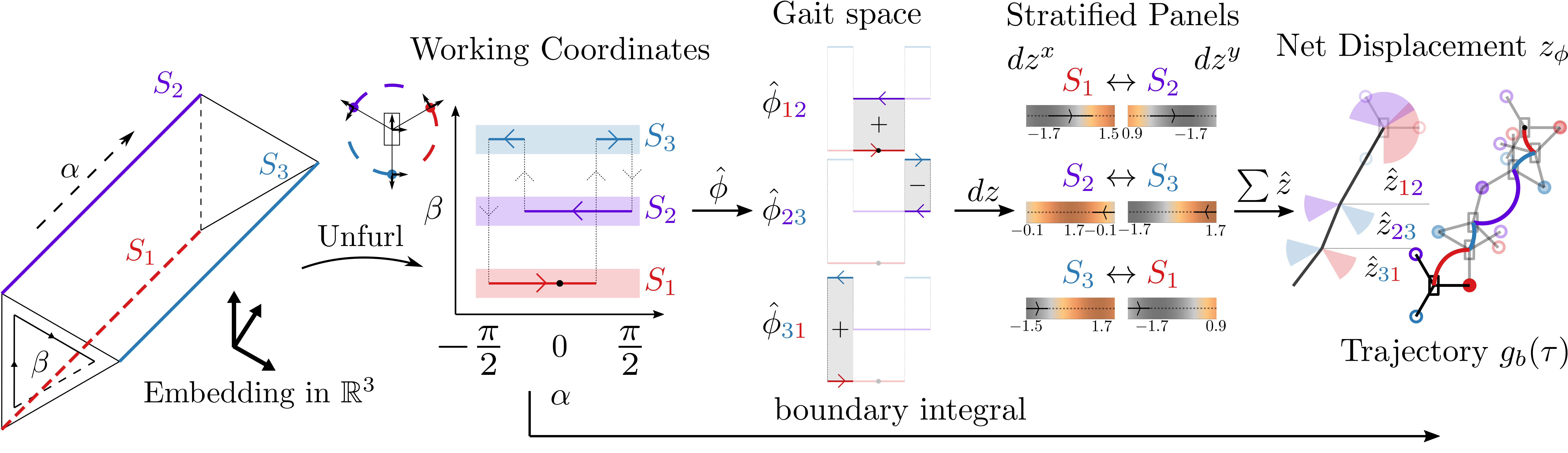}
    \caption{\hari{The natural $\mathbb{R}^3$ embedding of a three-footed system (inset) resembles a triangular prism due to a cyclic, continuous shape, $\alpha$, and a discrete shape $\beta$ encoding three locomotion submanifolds, $S_1$, $S_2$, and $S_3$. The embedding is opened along $\alpha$ and unfurled along $\beta$ to result in a shape-space similar to the two-footed system in \S \ref{sec:dual_cont}. From the working coordinates, the process of breaking down a multi-submanifold spanning, complex gait is shown sequentially from left to right. The gait is first broken down into sub-gaits in each panel and then each sub-gait is used to estimate the net displacement accrued by the system over a gait cycle. The swing ranges of each leg over a gait cycle are shown as highlighted circular sectors on the right.}}
    \label{fig:3C}
\end{figure*}

Following analysis in Sec. \ref{sec:dual_cont}, we derive the three local connections ($\boldsymbol{A}_{1}$, $\boldsymbol{A}_{2}$, and $\boldsymbol{A}_{3}$ during $S_{1}$, $S_{2}$ and $S_{3}$ respectively) and three stratified panels ($dz_{12}$, $dz_{23}$, and $dz_{31}$) representing the infinitesimal, contact-switching kinematics of our system \hari{(Fig. \ref{fig:3C})}. 

\vspace{-10pt}
{\subsection{Gait Coordinate System}}
\label{subsec:general}

Using stratified panels as a natural {gait coordinate system} \cite{hatton2011coordinatechoice} allows us to decompose a complex gait spanning multiple locomotion submanifolds into a sum of sub-gaits in each stratified panel for motion estimation. 
For this process of \textit{gait reduction} (and the reverse \textit{gait lifting}), counter-clockwise gaits are considered positive and $\hat{\phi}_{ij} \left(\alpha^{-}, \alpha^{+} \right)$ correspond to sub-gaits spanning the manifolds $S_i$ and $S_j$ with a swing range between $\alpha^{-}$ and $\alpha^{+}$. 
The robot's trajectory is obtained using the boundary integral as before using the knowledge of currently active submanifold as the gait progresses. 

To demonstrate the strength of our method, we choose an example complex gait ($\phi_{123}$) that spans all valid locomotion submanifolds of our three-footed system \hari{(Fig. \ref{fig:3C})} and decompose it into a gait-space representation (\hari{which may not be unique operations}) as:
\begin{equation} \label{eq:gait_basis_example}
    \phi_{123} \rightarrow \hat{\phi}_{12} \left(-\frac{\pi}{4}, \frac{\pi}{4} \right) - \hat{\phi}_{23} \left(\frac{\pi}{4}, \frac{\pi}{2} \right) + \hat{\phi}_{31} \left(-\frac{\pi}{2}, -\frac{\pi}{4} \right)
\end{equation}

%\vspace{-10pt}
% \section{Discussion and Future Work}
\section{\hari{Discussion}}
\label{sec:disc}

To summarize, in this work, we modeled the no-slip contact-switching quasi-static locomotion from a geometric mechanics perspective using the simplest instantiation of a hybrid system --- a two-footed toy robot with one continuous limb swing shape variable and one discrete foot contact shape variable, a departure from previous work \cite{chong2022coordinatingtiny, chong2021coordbackbend, chong2019hierarchical}. 
Leveraging the piece-wise holonomic nature of walking \cite{zhao2022walking}, we derived unique local connections for each contact submanifold and proved that the curvature of connection vector fields for a contact-switching system (typically computed from generalized Stokes' theorem) is simply the difference between those local connections. 
These discrete-curvature functions called stratified panels encode the infinitesimal displacement strength in a hybrid shape-space and when integrated over a gait period estimate the average motion over that cycle. We also demonstrated the ability to optimize the above gaits based on user-defined locomotion constraints.
Furthermore, these stratified panels represent a natural gait coordinate system that can reduce high-dimensional complex gaits into lower-dimensional sub-gaits enabling us to potentially scale our approach to multilegged systems which we illustrated by extending our analysis to a three-footed system.
Thus, using the stratified panels, our work has extended the use of generalized Stokes' theorem \cite{hatton2015nonconservativity} to classes of systems with hybrid shape-spaces consisting of both continuous and discrete shapes such as animals and robots walking without slipping \cite{ruina1998nonholonomic}. 

While the current results are promising, we envision a number of improvements and exciting future directions and a few are listed below. Our two- and three-footed systems provided useful insights into the stratified nature of the constraint curvature, but do not account for the variety of morphologies in sprawled multilegged systems. For example, many robots and animals have distinct hip joints with independently articulated legs. As the next steps, we aim to extend the current results to multi-disk multilegged systems that more closely replicate real systems. With such a model, we wish to generate optimal gait trajectories and evaluate them experimentally. 

%\hari{State-of-the-art legged robotic systems \cite{hutter2016anymal, katz2019MITminiCheetah} have shown impressive feats such as agile running \cite{park2017highspeedMIT}, locomotion in man-made \cite{gehring2021fieldANYmal} and natural environments \cite{lee2020learningchallengingterrain}, and more recently running on deformable terrain \cite{choi2023learnQuadDeformTerrain}. Such capabilities are a direct reflection of their system-level design and control architecture, typically characterized by overactuation, large computational resources, high-bandwidth control, and motion planning modules. For systems in the palm-size\cite{birkmeyer2009dash} and smaller scales\cite{jayaram2020scalinghamr, mi2022omni}, that lack power and computational budget, there exists a general need for reduced order models (ROMs) as a target for control.} 

\hari{State-of-the-art legged robotic systems have recently demonstrated  impressive locomotion feats such as agile and robust locomotion over complex human-made and natural terrains, relying on computational intelligence from efficient onboard processing \cite{lee2020learningchallengingterrain}.
However, a majority of these techniques like model predictive control still remain intractable for insect-scale robots and this is where we envision major applications of our work.}
\hari{We believe future flavors of the geometric, \emph{average}
%\footnote{Refer to \emph{holonomy} in \cite{kelly1995geometric}.}
locomotion models (or geometric reduced order models) like the one presented in this paper will lay the foundation for building efficient motion planners requiring minimal computational \hari{resources, and are} especially attractive for miniature robotic systems \cite{jayaram2020hamrjr,doshi2019effectivelocomultifreq}. 
We envision such models are critical for enabling predictive controllers in miniature robots to enable them to approach performances similar to their larger counterparts \cite{ choi2023learnQuadDeformTerrain, lee2020learningchallengingterrain}.}
These robots can use odometry \cite{jayaram2018concomitant} to build coarse approximations of the stratified panels leading to online gait synthesis (or even geometric model discovery, by extension). Similarly, these methods can be used to generate data-driven, principally kinematic models of animal walking to provide new insights into biological locomotion \cite{zhao2022walking}, which can then inspire new robot designs with enhanced locomotion capabilities (e.g., omnidirectional maneuverability enabled by shape-morphing robots like CLARI \cite{heiko2023_CLARI_intro}).
\bibliographystyle{ieeetran}
\bibliography{IEEEabrv, geomToyProblem}

\begin{thebibliography}{10}
\providecommand{\url}[1]{#1}
\csname url@rmstyle\endcsname
\providecommand{\newblock}{\relax}
\providecommand{\bibinfo}[2]{#2}
\providecommand\BIBentrySTDinterwordspacing{\spaceskip=0pt\relax}
\providecommand\BIBentryALTinterwordstretchfactor{4}
\providecommand\BIBentryALTinterwordspacing{\spaceskip=\fontdimen2\font plus
\BIBentryALTinterwordstretchfactor\fontdimen3\font minus
  \fontdimen4\font\relax}
\providecommand\BIBforeignlanguage[2]{{%
\expandafter\ifx\csname l@#1\endcsname\relax
\typeout{** WARNING: IEEEtran.bst: No hyphenation pattern has been}%
\typeout{** loaded for the language `#1'. Using the pattern for}%
\typeout{** the default language instead.}%
\else
\language=\csname l@#1\endcsname
\fi
#2}}

\bibitem{kelly1995geometric}
S.~D. Kelly and R.~M. Murray, ``Geometric phases and robotic locomotion,''
  \emph{Journal of Robotic Systems}, vol.~12, no.~6, pp. 417--431, 1995.

\bibitem{oliva2004geometric}
W.~M. Oliva, \emph{Geometric mechanics}.\hskip 1em plus 0.5em minus 0.4em\relax
  Springer, 2004.

\bibitem{hatton2015nonconservativity}
R.~L. Hatton and H.~Choset, ``Nonconservativity and noncommutativity in
  locomotion,'' \emph{European Phy. J. Special Topics}, vol. 224, no.~17, pp.
  3141--3174, 2015.

\bibitem{bittner2018geometrically}
B.~Bittner, R.~L. Hatton, and S.~Revzen, ``Geometrically optimal gaits: a
  data-driven approach,'' \emph{Nonlinear Dynamics}, vol.~94, pp. 1933--1948,
  2018.

\bibitem{astley2020surprising}
H.~C. Astley, J.~R. Mendelson~III, J.~Dai, C.~Gong, B.~Chong, J.~M. Rieser,
  P.~E. Schiebel, S.~S. Sharpe, R.~L. Hatton, H.~Choset, \emph{et~al.},
  ``Surprising simplicities and syntheses in limbless self-propulsion in
  sand,'' \emph{J. Exp. Biology}, vol. 223, no.~5, p. jeb103564, 2020.

\bibitem{chong2022coordinatingtiny}
B.~Chong, T.~Wang, E.~Erickson, P.~J. Bergmann, and D.~I. Goldman,
  ``Coordinating tiny limbs and long bodies: Geometric mechanics of lizard
  terrestrial swimming,'' \emph{Proc. of the Nat. Acad. of Sci.}, vol. 119,
  no.~27, p. e2118456119, 2022.

\bibitem{jacobs2012geometric}
H.~O. Jacobs, ``Geometric descriptions of couplings in fluids and circuits,''
  Ph.D. dissertation, California Institute of Technology, 2012.

\bibitem{hatton2013geometric}
R.~L. Hatton and H.~Choset, ``Geometric swimming at low and high reynolds
  numbers,'' \emph{IEEE Trans. on Robotics}, vol.~29, no.~3, pp. 615--624,
  2013.

\bibitem{dai2016geometric}
J.~Dai, H.~Faraji, C.~Gong, R.~L. Hatton, D.~I. Goldman, and H.~Choset,
  ``Geometric swimming on a granular surface.'' in \emph{Robotics: Science and
  Systems}, 2016, pp. 1--7.

\bibitem{zhao2022walking}
D.~Zhao, B.~Bittner, G.~Clifton, N.~Gravish, and S.~Revzen, ``Walking is like
  slithering: A unifying, data-driven view of locomotion,'' \emph{Proc. of the
  Nat. Acad. of Sci.}, vol. 119, no.~37, p. e2113222119, 2022.

\bibitem{chong2021coordbackbend}
B.~Chong, Y.~O. Aydin, C.~Gong, G.~Sartoretti, Y.~Wu, J.~M. Rieser, H.~Xing,
  P.~E. Schiebel, J.~W. Rankin, K.~B. Michel, \emph{et~al.}, ``Coordination of
  lateral body bending and leg movements for sprawled posture quadrupedal
  locomotion,'' \emph{Int. J. of Robotics Research}, vol.~40, no. 4-5, pp.
  747--763, 2021.

\bibitem{chong2023frictionalswimming}
B.~Chong, J.~He, S.~Li, E.~Erickson, K.~Diaz, T.~Wang, D.~Soto, and D.~I.
  Goldman, ``Self-propulsion via slipping: Frictional swimming in multilegged
  locomotors,'' \emph{Proc. of the Nat. Acad. of Sci.}, vol. 120, no.~11, p.
  e2213698120, 2023.

\bibitem{chong2019hierarchical}
B.~Chong, Y.~Ozkan~Aydin, G.~Sartoretti, J.~M. Rieser, C.~Gong, H.~Xing,
  H.~Choset, and D.~I. Goldman, ``A hierarchical geometric framework to design
  locomotive gaits for highly articulated robots,'' in \emph{Robotics: science
  and systems}, 2019.

\bibitem{chong2022general}
B.~Chong, Y.~O. Aydin, J.~M. Rieser, G.~Sartoretti, T.~Wang, J.~Whitman,
  A.~Kaba, E.~Aydin, C.~McFarland, K.~D. Cruz, \emph{et~al.}, ``A general
  locomotion control framework for multi-legged locomotors,''
  \emph{Bioinspiration \& Biomimetics}, vol.~17, no.~4, p. 046015, 2022.

\bibitem{chong2021moving}
B.~Chong, T.~Wang, B.~Lin, S.~Li, H.~Choset, G.~Blekherman, and D.~Goldman,
  ``Moving sidewinding forward: optimizing contact patterns for limbless robots
  via geometric mechanics,'' in \emph{Robotics: science and systems}, vol.~17,
  2021.

\bibitem{chong2023optimizing}
B.~Chong, T.~Wang, L.~Bo, S.~Li, P.~C. Muthukrishnan, J.~He, D.~Irvine,
  H.~Choset, G.~Blekherman, and D.~I. Goldman, ``Optimizing contact patterns
  for robot locomotion via geometric mechanics,'' \emph{The International
  Journal of Robotics Research}, p. 02783649231188387, 2023.

\bibitem{hatton2022geometry}
R.~L. Hatton, Z.~Brock, S.~Chen, H.~Choset, H.~Faraji, R.~Fu, N.~Justus, and
  S.~Ramasamy, ``The geometry of optimal gaits for inertia-dominated kinematic
  systems,'' \emph{IEEE Transactions on Robotics}, vol.~38, no.~5, pp.
  3279--3299, 2022.

\bibitem{goodwine2002motionkinstrat}
B.~Goodwine and J.~Burdick, ``Motion planning for kinematic stratified systems
  with application to quasi-static legged locomotion and finger gaiting,''
  \emph{IEEE Trans. on Rob. \& Automation}, vol.~18, no.~2, pp. 209--222, 2002.

\bibitem{de2018inverted}
S.~D. De~Rivaz, B.~Goldberg, N.~Doshi, K.~Jayaram, J.~Zhou, and R.~J. Wood,
  ``Inverted and vertical climbing of a quadrupedal microrobot using
  electroadhesion,'' \emph{Science Robotics}, vol.~3, no.~25, p. eaau3038,
  2018.

\bibitem{ruina1998nonholonomic}
A.~Ruina, ``Nonholonomic stability aspects of piecewise holonomic systems,''
  \emph{Reports on mathematical physics}, vol.~42, no. 1-2, pp. 91--100, 1998.

\bibitem{full1999templates}
R.~J. Full and D.~E. Koditschek, ``Templates and anchors: neuromechanical
  hypotheses of legged locomotion on land,'' \emph{Journal of experimental
  biology}, vol. 202, no.~23, pp. 3325--3332, 1999.

\bibitem{hatton2017Riemannian}
R.~L. Hatton, T.~Dear, and H.~Choset, ``Kinematic cartography and the
  efficiency of viscous swimming,'' \emph{IEEE Trans. on Robotics}, vol.~33,
  no.~3, pp. 523--535, 2017.

\bibitem{hatton2013geometric2}
R.~L. Hatton, Y.~Ding, H.~Choset, and D.~I. Goldman, ``Geometric visualization
  of self-propulsion in a complex medium,'' \emph{Physical review letters},
  vol. 110, no.~7, p. 078101, 2013.

\bibitem{gray1955propulsion}
J.~Gray and G.~Hancock, ``The propulsion of sea-urchin spermatozoa,''
  \emph{Journal of Experimental Biology}, vol.~32, no.~4, pp. 802--814, 1955.

\bibitem{zhao2021locomotion}
D.~Zhao, ``Locomotion of low-dof multi-legged robots,'' Ph.D. dissertation, Ph.
  D. dissertation, University of Michigan, 2021.

\bibitem{wu2019coulomb}
Z.~Wu, D.~Zhao, and S.~Revzen, ``Coulomb friction crawling model yields linear
  force--velocity profile,'' \emph{Journal of Applied Mechanics}, vol.~86,
  no.~5, p. 054501, 2019.

\bibitem{zhang2014effectiveness}
T.~Zhang and D.~I. Goldman, ``The effectiveness of resistive force theory in
  granular locomotion,'' \emph{Physics of Fluids}, vol.~26, no.~10, 2014.

\bibitem{ramasamy2019draggeometry}
S.~Ramasamy and R.~L. Hatton, ``The geometry of optimal gaits for
  drag-dominated kinematic systems,'' \emph{IEEE Trans. on Robotics}, vol.~35,
  no.~4, pp. 1014--1033, 2019.

\bibitem{hatton2011coordinatechoice}
R.~L. Hatton and H.~Choset, ``Geometric motion planning: The local connection,
  stokes’ theorem, and the importance of coordinate choice,'' \emph{Int. J.
  of Robotics Research}, vol.~30, no.~8, pp. 988--1014, 2011.

\bibitem{lee2020learningchallengingterrain}
J.~Lee, J.~Hwangbo, L.~Wellhausen, V.~Koltun, and M.~Hutter, ``Learning
  quadrupedal locomotion over challenging terrain,'' \emph{Science robotics},
  vol.~5, no.~47, p. eabc5986, 2020.

\bibitem{jayaram2020hamrjr}
K.~Jayaram, J.~Shum, S.~Castellanos, E.~F. Helbling, and R.~J. Wood, ``Scaling
  down an insect-size microrobot, hamr-vi into hamr-jr,'' in \emph{2020 IEEE
  International Conference on Robotics and Automation (ICRA)}.\hskip 1em plus
  0.5em minus 0.4em\relax IEEE, 2020, pp. 10\,305--10\,311.

\bibitem{doshi2019effectivelocomultifreq}
N.~Doshi, K.~Jayaram, S.~Castellanos, S.~Kuindersma, and R.~J. Wood,
  ``Effective locomotion at multiple stride frequencies using proprioceptive
  feedback on a legged microrobot,'' \emph{Bioinspiration \& biomimetics},
  vol.~14, no.~5, p. 056001, 2019.

\bibitem{choi2023learnQuadDeformTerrain}
S.~Choi, G.~Ji, J.~Park, H.~Kim, J.~Mun, J.~H. Lee, and J.~Hwangbo, ``Learning
  quadrupedal locomotion on deformable terrain,'' \emph{Science Robotics},
  vol.~8, no.~74, p. eade2256, 2023.

\bibitem{jayaram2018concomitant}
K.~Jayaram, N.~T. Jafferis, N.~Doshi, B.~Goldberg, and R.~J. Wood,
  ``Concomitant sensing and actuation for piezoelectric microrobots,''
  \emph{Smart Materials and Structures}, vol.~27, no.~6, p. 065028, 2018.

\bibitem{heiko2023_CLARI_intro}
\BIBentryALTinterwordspacing
H.~Kabutz and K.~Jayaram, ``Design of clari: A miniature modular origami
  passive shape-morphing robot,'' \emph{Advanced Intelligent Systems}, vol.
  n/a, no. n/a, p. 2300181, 2023. [Online]. Available:
  \url{https://onlinelibrary.wiley.com/doi/abs/10.1002/aisy.202300181}
\BIBentrySTDinterwordspacing

\end{thebibliography}

\end{document}